\pdfoutput=1

\documentclass[11pt]{article}

\usepackage[]{acl}

\usepackage{times}
\usepackage{latexsym}

\usepackage[T1]{fontenc}

\usepackage[utf8]{inputenc}

\usepackage{microtype}
\usepackage{natbib}
\usepackage{booktabs}
\usepackage{graphicx}
\usepackage{multirow}
\usepackage{ulem}
\usepackage{float}
\usepackage{subfigure}
\usepackage{multirow}
\usepackage{pgfplots}
\usepackage{amsmath}
\usepackage{svg}
\usepackage{hyperref}

\title{FSUIE: A Novel Fuzzy Span Mechanism for Universal Information Extraction}

\author{Tianshuo Peng$^{1,\dag}$, Zuchao Li$^{1,2,\dag,}$\thanks{$\ $  Corresponding author. $^\dag$ Equal contribution. This work was supported by the Fundamental Research Funds for the Central Universities (No. 2042023kf0133), the Special Fund of Hubei Luojia Laboratory under Grant 220100014 and the National Science Fund for Distinguished Young Scholars under Grant 62225113. Hai Zhao was funded by the Key Projects of National Natural Science Foundation of China (U1836222 and 61733011).}, Lefei Zhang$^{1,2}$, Bo Du$^{1,2}$ and Hai Zhao$^{3}$\\
$^{1}$National Engineering Research Center for Multimedia Software, \\
School of Computer Science, Wuhan University, Wuhan, 430072, P. R. China \\
$^{2}$Hubei Luojia Laboratory, Wuhan 430072, P. R. China \\
$^{3}$Department of Computer Science and Engineering, Shanghai Jiao Tong University\\
{\tt \{pengts,zcli-charlie,zhanglefei,dubo\}@whu.edu.cn,}\\
{\tt zhaohai@cs.sjtu.edu.cn}\\
}

\begin{document}
\maketitle

\begin{abstract}

Universal Information Extraction (UIE) has been introduced as a unified framework for various Information Extraction (IE) tasks and has achieved widespread success. Despite this, UIE models have limitations. For example, they rely heavily on span boundaries in the data during training, which does not reflect the reality of span annotation challenges. Slight adjustments to positions can also meet requirements. Additionally, UIE models lack attention to the limited span length feature in IE. To address these deficiencies, we propose the Fuzzy Span Universal Information Extraction (FSUIE) framework. Specifically, our contribution consists of two concepts: \textit{fuzzy span loss} and \textit{fuzzy span attention}. Our experimental results on a series of main IE tasks show significant improvement compared to the baseline, especially in terms of fast convergence and strong performance with small amounts of data and training epochs. These results demonstrate the effectiveness and generalization of FSUIE in different tasks, settings, and scenarios.

\end{abstract}

\section{Introduction}

Information Extraction (IE) is focused on extracting predefined types of information from unstructured text sources, such as Named Entity Recognition (NER), Relationship Extraction (RE), and Sentiment Extraction (SE). To uniformly model the various IE tasks under a unified framework, a generative Universal Information Extraction (UIE) was proposed in \cite{lu-etal-2022-unified} and has achieved widespread success on various IE datasets and  benchmarks. Due to the necessity of a powerful generative pre-training model for the generative UIE, the time overhead is extensive and the efficiency is not satisfactory. For this reason, this paper examines span-based UIE to unify various IE tasks, conceptualizing IE tasks as predictions of spans.

\begin{figure}[t]
    \centering
    \includegraphics[width=0.5\textwidth]{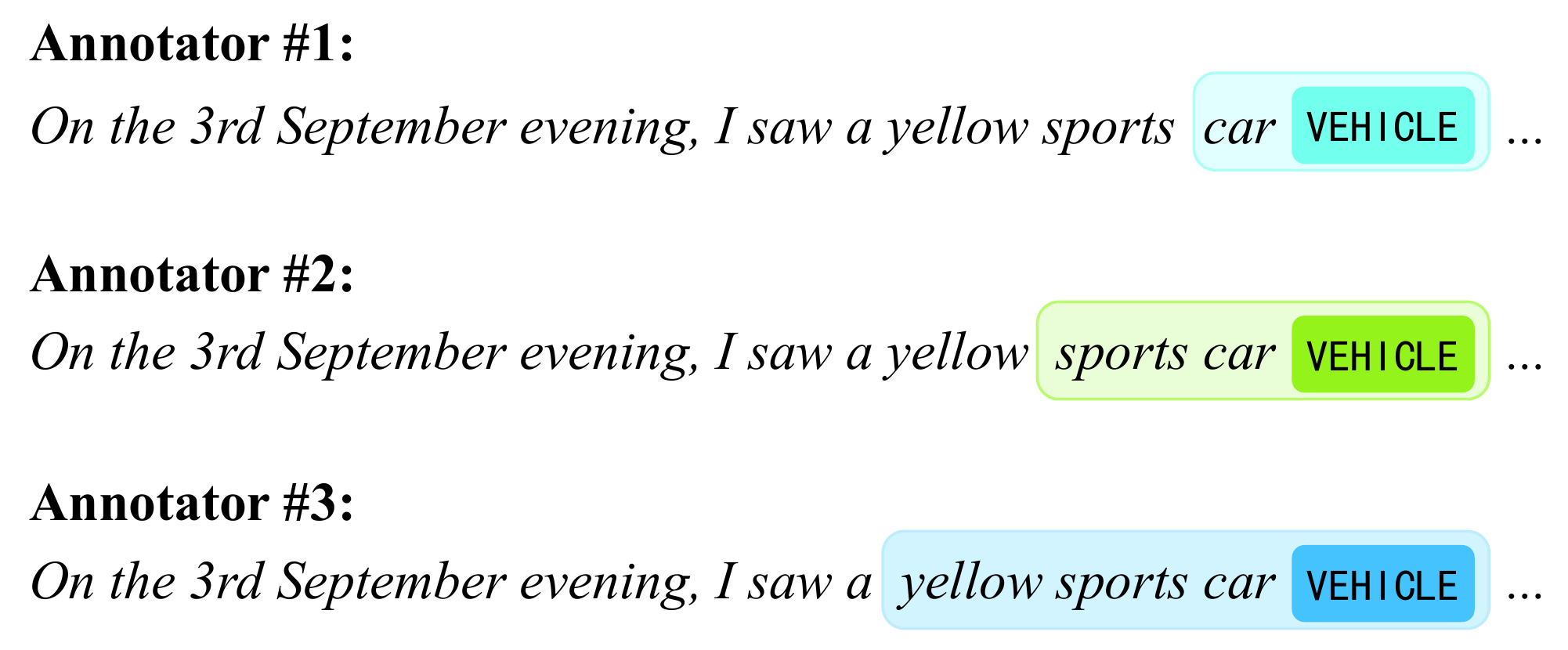}
    \caption{An example of annotation ambiguity for ``vehicle" entity in sentence ``\textit{On the 3rd September evening, I saw a yellow sports car drive past my house}".}%
    \label{fig:fuzzy_boundary}
\end{figure}

However, UIE models still have some limitations. First, as it is the process of training machine learning models to extract specific information from unstructured text sources, IE relies heavily on human annotation which involves labeling the data by identifying the specific information to be extracted and marking the corresponding span boundaries in the text manually. However, due to the complexity of natural language, determining the correct span boundaries can be challenging, leading to the phenomenon of annotation ambiguity. As shown in Figure \ref{fig:fuzzy_boundary}, different annotated spans can be considered reasonable. In the span learning of UIE models, the method of teacher forcing is commonly used for loss calculation, making the model dependent on the precise span boundaries given in the training data. This can cause performance bottlenecks due to annotation ambiguity.

When the model structure in UIE places too much emphasis on the exact boundaries of IE tasks, it leads to insufficient utilization of supervision information. In order to predict span boundaries, positions closer to the ground-truth should be more accurate than those relatively farther away, as shown in Figure \ref{fig:fuzzy_boundary}. For example, words close to the target “car” are more likely to be correct than the word “evening” which is farther away from the target. Under the premise of positioning to the span where "car" is located, both the "yellow car" and the "yellow sports car" can be regarded as vehicle entities. This means that the span model learned should be fuzzy rather than precise.

In addition, the use of pre-trained Transformer~\cite{DBLP:conf/nips/VaswaniSPUJGKP17} in UIE to extract the start and end position representations also poses a problem. The Transformer model is designed to focus on the global representation of the input text, while UIE requires focusing on specific parts of the text to determine the span boundaries. This mismatch between the Transformer's focus on global representation and UIE's focus on specific parts of the text can negatively impact the performance of the model.

When there is a mismatch between the Transformer architecture and the span representation learning, the model may not make good use of prior knowledge in IE. Specifically, given the start boundary (end boundary) of the label span, the corresponding end boundary (start boundary) is more likely to be found within a certain range before and after, rather than throughout the entire sequence. This is a prior hypotheses that span has limited length, which is ignored in the vanilla UIE model. To address this, a fuzzy span attention mechanism, rather than fixed attention, should be applied.

In this paper, we propose the Fuzzy Span Universal Information Extraction (FSUIE) framework that addresses the limitations of UIE models by applying the fuzzy span feature, reducing over-reliance on label span boundaries and adaptively adjusting attention span length. Specifically, to solve the issue of fuzzy boundaries, we design the \textit{fuzzy span loss} that quantitatively represents the correctness information distributed on fuzzy span. At the same time, we introduce \textit{fuzzy span attention} that sets the scope of attention to a fuzzy range and adaptively adjusts the length of span according to the encoding.
We conduct experiments on various main IE tasks (NER, RE, and ASTE). The results show that our FSUIE has a significant improvement compared to the strong UIE baseline in different settings. Additionally, it achieves new state-of-the-art performance on some NER, RE, and ASTE benchmarks with only bert-base architecture, outperforming models with stronger pre-trained language models and complex neural designs. Furthermore, our model shows extremely fast convergence, and good generalization on low-resource settings. These experiments demonstrate the effectiveness and generalization of FSUIE in different tasks, settings, and scenarios.

\section{FSUIE}

In FSUIE, incorporating fuzzy span into base UIE model involves two aspects. Firstly, for the spans carrying specific semantic types in the training data, the boundary targets should be learned as fuzzy boundaries to reduce over-reliance on span boundaries. To achieve this, we propose a novel \textit{fuzzy span loss}. Secondly, during the span representation learning, the attention applied in span should be dynamic and of limited length, rather than covering the entire sequence. To achieve this, we propose a novel \textit{fuzzy span attention}.

\subsection{Fuzzy Span Loss (FSL)}

The introduction of FSL is a supplement to traditional teacher forcing loss (usually implemented as Cross Entrophy), to guide the model in learning fuzzy boundaries. The challenge for FSL is how to quantify the distribution of correctness information within the fuzzy boundary. Specifically, for a given label span $S$, conventional target distributions (one-hot) indicate the correct starting and ending boundaries. This form actually follows the Dirac delta distribution that only focuses on the ground-truth positions, but cannot model the ambiguity phenomenon in boundaries.

To address the challenge discussed above, we propose a fuzzy span distribution generator (FSDG). In our method, we use a probability distribution of span boundaries to represent the ground-truth, which is more comprehensive in describing the uncertainty of boundary localization. It consists of two main steps: 1) determining the probability density distribution function $f$; 2) mapping from the continuous distribution to a discrete probability distribution based on $f$.

Specifically, let $q \in S$ be a boundary of the label span, then the total probability value of its corresponding fuzzy boundary $\hat{q}$ can be represented as follows:
\begin{equation}
   \hat{q}=\int_{R_{min}}^{R_{max}} x Q(x) d x, \quad q \in S
   \label{eq:p}
\end{equation}
where $x$ represents the coordinate of boundaries within the fuzzy range $[R_{min} , R_{max} ]$, $R_{min}$ and $R_{max}$ are the start and end positions of the fuzzy range, $q^{gt}$ represents the ground-truth position for boundary $q$, and $Q(x)$ represents the corresponding coordinate probability.

\begin{figure}[t]
    \centering
    \includegraphics[width=0.5\textwidth]{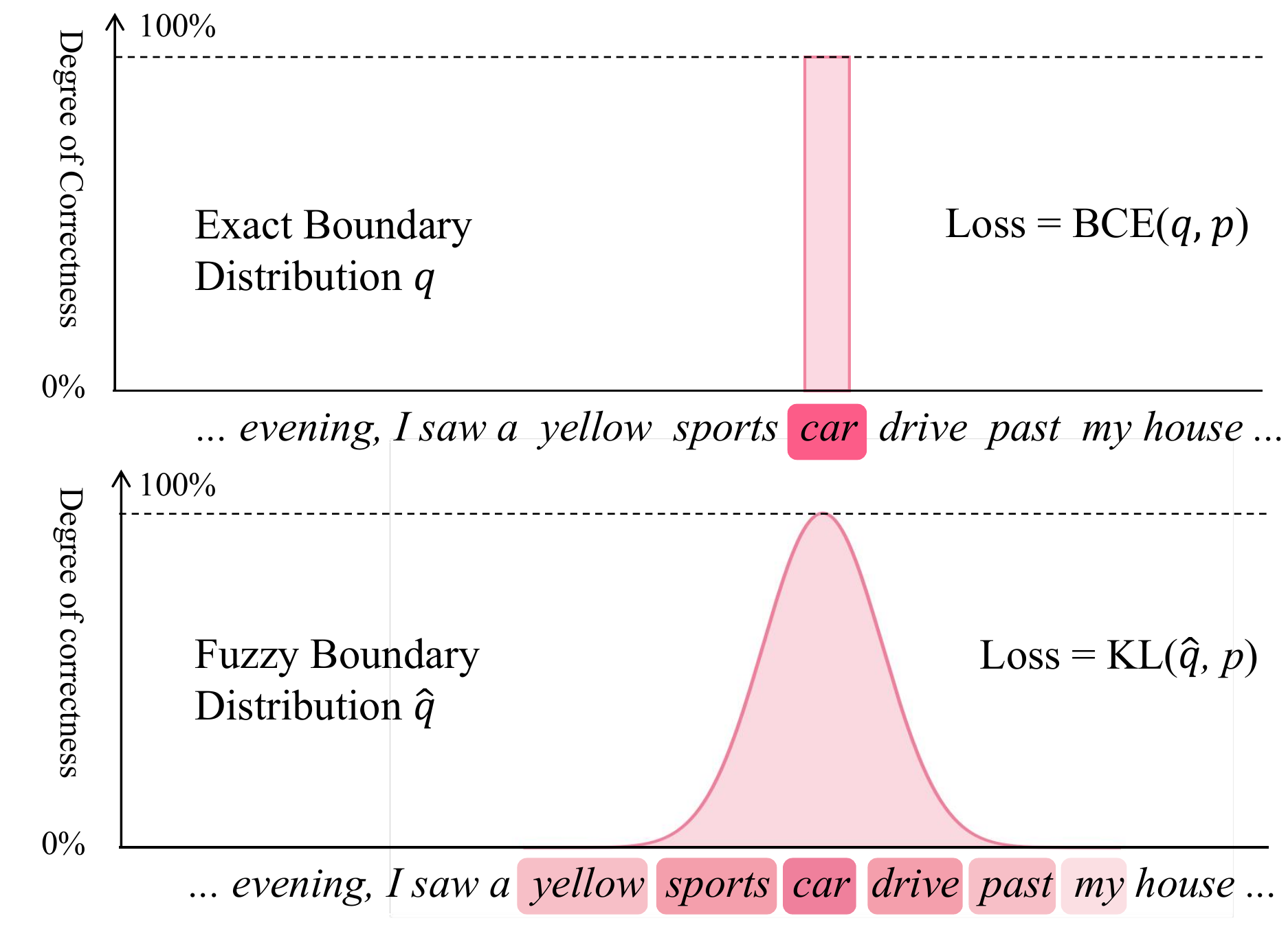}
    \vspace{-10pt}
    \caption{Illustration of exact boundary and fuzzy boundary.}
    \label{fig:fsl_illu}
\end{figure}

The traditional Dirac delta distribution can be viewed as a special case of Eq. (\ref{eq:p}), where $Q(x) = 1$ when $x = q^{gt}$, and $Q(x) = 0$ otherwise. Through a mapping function $F$, we can quantifying continuous fuzzy boundaries into a unified discrete variable 
$\mathbf{\hat{q}}=[F(q_{1}), F(q_{2}), \cdots, F(q_{n})]$
with $n$ subintervals. $[q_{1}, q_{2}, \cdots, q_{n}]$ represent continuous coordinates in fuzzy range where $q_{1}=R_{min}$ and $q_{n}=R_{max}$, the probability distribution of each given boundary of the label span can be represented within the range via the softmax function.

Since the Dirac delta distribution only assigns non-zero probability to a single point, it is not suitable for modeling uncertainty or ambiguity in real-world data. Thus in FSUIE, we choose the Gaussian distribution $N(\mu,\sigma^{2})$ as the probability density function $f$. 
Compared with other probability distributions, the Gaussian distribution assigns non-zero probability to an entire range of values has the following advantages: (1) it is continuous, symmetrical, and can well represent the distribution of correctness information within the fuzzy boundary including the gold position; (2) it is a stable distribution with fewer peaks and offsets, and can ensure that the correctness information is more concentrated on the gold position while distributed on the fuzzy boundary; (3) the integral of the Gaussian distribution is 1, which can ensure that the accuracy distribution after softmax is more gentle.

To get the discrete variable 
$\mathbf{\hat{q}}$
, Four parameters are involved here: variance $\sigma$, mean $\mu$, sampling step $s$, and sampling threshold $\theta$. These parameters are used to control the range, peak position, and density of the fuzzy boundary, respectively. Specifically,  the parameter $\mu$ is set to $q^{gt}$ and the Gaussian distribution is determined using a pre-determined $\sigma$. Assuming $q_{g}\in [q_{1}, q_{2}, \cdots, q_{n}] =q^{gt}$, $F$ can represented as:
\begin{equation}
    \begin{aligned}
    &F(q_{i})=\left\{\begin{array}{ll}
    \varepsilon, &\varepsilon \ge \theta \\
    0, &\varepsilon <\theta \\ 
    \end{array}\right.,\\
    &\varepsilon=f(\mu+(i-g)s).
    \end{aligned}
\end{equation}

Given that values in the marginal regions of Gaussian distribution are quite small, the sampling threshold $\theta$ here acts as a filter to eliminate information from unimportant locations.
The specific choice of parameters is discussed in the following experimental section.
We use 
$\mathbf{\hat{q}}$
as the distribution of correctness information on the fuzzy boundaries. The beginning and end fuzzy boundaries together make up the fuzzy span. Then, we calculate the KL divergence between the model's predicted logits and the gold fuzzy span distribution as the \textit{fuzzy span loss}. The exact boundary and fuzzy boundary distribution is shown in Figure \ref{fig:fsl_illu}. This \textit{fuzzy span loss} is then incorporated into the original teacher-forcing loss function with a coefficient:

\begin{equation*}
    \begin{aligned}
    \mathcal{L}_{FS}=D_{K L}(\mathbf{\hat{q}} \| p)&=\sum_{i=1}^{N} \mathbf{\hat{q}}\left(x_{i}\right)\left(\log \frac{\mathbf{\hat{q}}\left(x_{i}\right)}{p\left(x_{i}\right)}\right), \\
    \mathcal{L}&=\mathcal{L}_{ori}+\lambda \mathcal{L}_{FS} \\
    \end{aligned}
\end{equation*}

where $p$ represents the predicted distribution of the model and
$\mathbf{\hat{q}}$
represents the generated fuzzy span distribution from FSDG according to the annotation in training data. $\mathcal{L}_{ori}$ is the original Binary Cross Entropy (BCE) loss of the model in UIE, and $\lambda$ is the coefficient of the \textit{fuzzy span loss}.

\subsection{Fuzzy Span Attention (FSA)}

We construct a FSA based on a multi-head self-attention mechanism with relative positional encoding (RPE), since RPE is more suitable for span representation learning with fuzzy bounds. In conventional multi-head attention with RPE, for a token at position $t$ in the sequence, each head computes the similarity matrix of this token and the tokens in the sequence. The similarity between token $t$ and token $r$ can be represented as:
\begin{equation}
s_{t r}=y_{t}^{\top} W_{q}^{\top}\left(W_{k} y_{r}+p_{t-r}\right)
\end{equation}
where $W_k$ and $W_q$ are the weight matrices for "key" and "query" representations, $y_t$ and $y_r$ are the representations of token $t$ and $r$, and $p_{t-r}$ is the relative position embedding, the corresponding attention weight can be obtained through a softmax function
\begin{equation}
    a_{t r}=\frac{\exp \left(s_{t r}\right)}{\sum_{q=0}^{t-1} \exp \left(s_{t q}\right)}.
\end{equation}

Conventional self-attention focus on global representations, mismatching the requirement of fuzzy spans. To address this issue, we present a novel attention mechanism, called \textit{Fuzzy Span Attention} (FSA), to control attention scores of each token, aiming to learn a span-aware representation. The fuzzy span mechanism of FSA consists of two aspects: (1) the length of the range applying full attention is dynamically adjusted; and (2) the attention weights on the boundary of the full attention span are attenuating rather than truncated. Specifically, inspired by~\cite{sukhbaatar-etal-2019-adaptive}, we design a mask function $g_{m}$ to control the attention score calculation. Assuming the maximum length of the possible attention span is $L_{span}$, the new attention scores can be represented as:
\begin{equation}
    a_{t r}=\frac{g_{m}(t-r)\exp \left(s_{t r}\right)}{\sum_{q=t-L_{span}}^{t-1} g_{m}(t-r)\exp \left(s_{t q}\right)}.
\end{equation}
The following process is divided into two stages: (1) determining the attention changing function $g_{a}$ on the fuzzy span, and (2) constructing the mask function $g_{m}$ based on $g_{a}$ for span-aware representation learning.
According to the characteristics of fuzzy span, we set $g_{a}$ as a monotonically decreasing linear function. To adjust the attention span length, we define a learnable parameter $\delta \in [0,1]$. The $g_{a}(x)$ and corresponding $g_{m}(x)$ can be represented as follows:
\begin{equation}
\begin{aligned}
    g_{a}(z)&=\frac{-z+l+d}{d}, \\
    l&=\delta L_{span}.
\end{aligned}
\end{equation}
\begin{equation}
g_{m}(z)=\left\{\begin{array}{ll}
1, & g_{a}(z)>1 \\
0, & g_{a}(z)<0 \\
g_{a}(z), &  otherwise 
\end{array}\right..
\end{equation}
where $l$ controls the length of the full attention range and $d$ is a hyper-parameter that governs the length of the attenuated attention range.

\begin{figure} 
    \centering    
    \includegraphics[width=1\columnwidth]{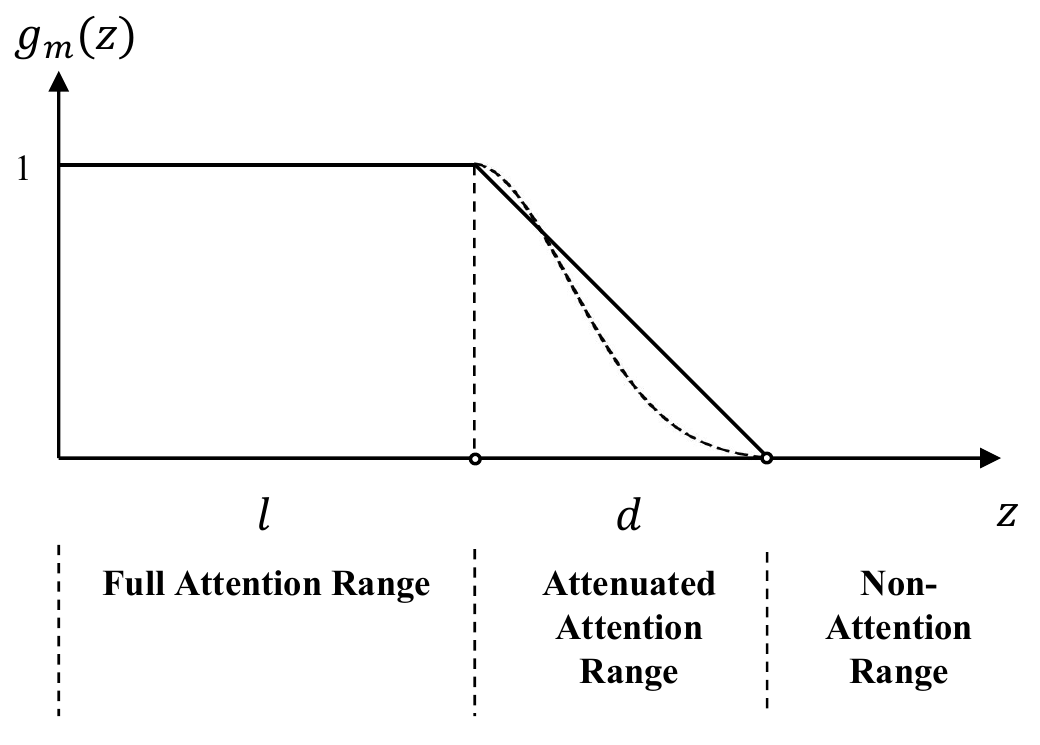}  
    \vspace{-10pt}
    \caption{Illustration of attention mask function $g_{m}$.}     
    \label{fig:fsa_illu}     
\end{figure}

In Figure~\ref{fig:fsa_illu}, an illustration of the $g_m$ function is depicted. The dashed lines represent alternative choices of $g_a$ functions, such as
\begin{equation*}
    g_{a}'(z)=\left\{\begin{array}{ll}
    1, & z \le l \\
    0, & z > l \\
\end{array}\right.,
\end{equation*}
\begin{equation*}
    g_{a}''(z)=\left\{\begin{array}{ll}
    1, & z \le l \\
    \frac{1}{\sqrt{2 \pi} \cdot \frac{d}{3} } \exp \left(-\frac{(z-l)^{2}}{2 (\frac{d}{3})^{2}}\right), & z > l \\
\end{array}\right..
\end{equation*}
Through experimentation, we found that the linear attenuated function performs best (refer to comparison in Appendix A). Iterative optimization of $\delta$ allows the model to learn the optimal attention span lengths for a specific task. It is important to note that different heads learn the attention span length independently and thus obtain different optimal fuzzy spans.
In our implementation, instead of using multiple layers of fuzzy span attention layers, we construct the span-aware representation with a single fuzzy span attention layer on top of Transformer encoder, and it does not participate in the encoding process.
Therefore, although the maximum range of fuzzy span attention is limited by $L_{span}$, it only affects span decisions and does not have any impact on the representation of tokens in the sequence.

\begin{table*}[t]
    \centering
    \small
    \resizebox{\textwidth}{!}{
    \begin{tabular}{lcccccccccc}
        \toprule
         \multirow{2}{*}{\bf Models} & \multirow{2}{*}{\bf PLM} & \multicolumn{3}{c}{\bf ACE04} & \multicolumn{3}{c}{\bf ACE05} & \multicolumn{3}{c}{\bf ADE}\\
         \cmidrule(lr){3-5} \cmidrule(lr){6-8} \cmidrule(lr){9-11} & & P & R & F$_1$ & P & R & F$_1$ & P & R & F$_1$ \\
         \midrule
         BENSC~\cite{DBLP:conf/aaai/TanQCWH20} & BERT-base & 85.80 & 84.80 & 85.30  & 83.80 & 83.90 & 83.90 & - & - & - \\ 
        \cite{DBLP:journals/corr/abs-2010-07522} & BERT-large  & - & - & - & - & - & 88.60 & - & - & - \\
        SpERT~\cite{DBLP:journals/corr/abs-1909-07755} & BERT-base  & - & - & - & - & - & - & 88.69 & 89.20 & 88.95 \\
        SpERT.PL~\cite{santosh2021joint} & Bio-BERT  & - & - & - & - & - & - & 90.05 & 91.69 & 90.86 \\
        BoningKnife~\cite{DBLP:journals/corr/abs-2107-09429} & BERT-base  & 85.98 & 86.86 & 86.41 & 84.77 & 86.16 & 85.46 & - & - & - \\ 
        JCBIE~\cite{he2022jcbie} & Bio-BERT & - & - & - & - & - & -  & - & - & 87.80  \\
        GLOBAL POINTER~\cite{DBLP:journals/corr/abs-2208-03054} & BERT-base & - & - & - & - & - & -  & - & - & 90.10 \\
        BS~\cite{zhu-li-2022-boundary} & RoBERTa-base & 88.43 & 87.53 & 87.98 & 86.25 & 88.07 & 87.15  & -  & -  & - \\
        Triaffine~\cite{yuan-etal-2022-fusing} & BERT-large & 87.13 & 87.68 & 87.40 & 86.70 & 86.94 & 86.82  & - & - & - \\
        Triaffine~\cite{yuan-etal-2022-fusing} & ALBERT-xxlarge   & 88.88 & 88.24 & \textbf{88.56} & 87.39 & 90.31 & \textbf{88.83} & - & -  & - \\
         \midrule
         \cite{yan-etal-2021-unified-generative} & BERT-large & 87.27 & 86.41 & 86.84 & 83.16 & 86.38 & 84.74  & - & - & - \\
         Generative UIE (SEL only)~\cite{lu-etal-2022-unified} & T5-v1.1-large  & -  & -  & 86.52 & -  & - & 85.52  & -  & - & - \\
         Generative UIE~\cite{lu-etal-2022-unified} & T5-v1.1-large   & -   & -  & 86.89 & -  & - & 85.78  & -   & -  & -  \\
         \midrule
         UIE-base & BERT-base & 88.25  & 80.31 & 84.09  & 86.19 & 83.12 & 84.63  & 87.85 & 91.56 & 89.67 \\
        \bf FSUIE-base  & BERT-base & 85.67 & 84.82 & 85.24 & 87.05 & 85.40 & 86.22  & 91.17 & 92.17 & 92.49 \\
        \bf FSUIE-large & BERT-large   &86.15 &86.17 &86.16 &88.06 &85.79 &86.91 &93.82 &92.21 &\textbf{93.08} \\
        \bottomrule
    \end{tabular}
    }
    \caption{NER experimental results on ACE04, ACE05, and ADE datasets.}
    \label{tab:ner_results}
\end{table*}

\section{Experiments}

\subsection{Setup}

\paragraph{Tasks} 

We conducted experiments on 4 datasets for 3 common information extraction tasks: NER, RE, and ASTE. The datasets we used include ACE2004, ACE2005, ADE~\cite{DBLP:journals/jbi/GurulingappaRRFHT12} for NER and RE tasks, and ASTE-Data-V2~\cite{xu-etal-2020-position} for ASTE task. 
We evaluate our model using different metrics for the three IE tasks. For NER, we use the Entity F$_1$ score, in which an entity prediction is correct if its span and type match a reference entity. For RE, we use the Relation Strict F$_1$ score, where a relation is considered correct only if its relation type and related entity spans are all correct. For ASTE, we use the Sentiment Triplet F$_1$ score, where a triplet is considered correct if the aspect, opinion, and sentiment polarity are all correctly identified.

\paragraph{Training Details} 

We trained two variations of FSUIE, FSUIE-base and FSUIE-large, which are based on the BERT-base and BERT-large model architecture and pre-training parameters respectively. In addition, we also trained a UIE-base based on BERT-base as a baseline without using FSL and FSA layers. In FSUIE, we added the FSA layer and the span boundary prediction layer to both models. Specifically, FSUIE-base has 12 layers of 12-head Transformer layers, with a hidden size of 768, while FSUIE-large has 24 layers of 16-head Transformer layers, with a hidden size of 1024. 
During training, we set the parameters of the Gaussian distribution in FSL as $\sigma=0.5$, the distribution value truncation threshold $\theta$ to 0.3, sampling step $s$ to 0.3, and the loss coefficient $\lambda$ to 0.01. And the parameter $\mu$ is set to the coordinate of annotation boundary.
The hyper-parameters $L_{span}$ and $d$ involved are determined based on the statistics of the target length on the UIE training data. During training, we set $L_{span}$ to $30$ and $d$ to $32$, and experimentation results have shown that the model's performance is not significantly sensitive to the choice of these hyper-parameters (refer to comparison in Appendix C).

We trained both models for 50 epochs with a learning rate of 1e-5 on the datasets of each task, and selected the final model based on the performance on the development set.
The code is available at
\href{https://github.com/pengts/FSUIE}{\tt https://github.com/pengts/FSUIE}.

\begin{table*}[t]
    \centering
    \small
    \resizebox{\textwidth}{!}{
    \begin{tabular}{lcccccccccc}
        \toprule
         \multirow{2}{*}{\bf Models} & \multirow{2}{*}{\bf PLM} & \multicolumn{3}{c}{\bf ACE04} & \multicolumn{3}{c}{\bf ACE05} & \multicolumn{3}{c}{\bf ADE}\\
         \cmidrule(lr){3-5} \cmidrule(lr){6-8} \cmidrule(lr){9-11} & & P & R & F$_1$ & P & R & F$_1$ & P & R & F$_1$ \\
         \midrule
         SpERT~\cite{DBLP:journals/corr/abs-1909-07755} & BERT-base & - & -  & - & -  & -  & -   & 77.77 & 79.96 & 78.84 \\
         SpERT.PL~\cite{santosh2021joint}  & Bio-BERT   & -  & - & - & - & -  & -   & 80.11       & 84.18 & 82.03 \\
        JCBIE~\cite{he2022jcbie} & Bio-BERT & - & - & - & - & - & -  & - & - & 74.18 \\
        \cite{DBLP:journals/corr/abs-2010-07522}  & BERT-base & - & - & - & -  & -  & 66.10 & - 
        & - & - \\
        \cite{DBLP:journals/corr/abs-2010-07522} & BERT-large & - & -  & - & - & - & 68.10  & -       & -  & - \\
        Tabel-Sequence Encoder~\cite{wang-lu-2020-two} & ALBERT-xxlarge & - & - & 63.30 & - & - & 67.60  & -  & - & 80.10 \\ 
        PL-Marker~\cite{ye-etal-2022-packed} & BERT-base & - & - & 66.70  & -  & - & 69.00  & -   & - & -   \\
        PL-Marker~\cite{ye-etal-2022-packed} & ALBERT-xxlarge   & -  & - & 69.70 & -  & - & 73.00   & - & - & -  \\
        \midrule
        Generative UIE (SEL only)~\cite{lu-etal-2022-unified} & T5-v1.1-large & - & - & - & - & - & 64.68  & - & - & - \\
        Generative UIE~\cite{lu-etal-2022-unified} & T5-v1.1-large   & - & - & - & - & - & 66.06  & - & - & - \\
        \midrule
        UIE-base & BERT-base   & 88.73 & 53.46 & 66.72 &95.38  &52.04  &67.34 & 67.61 & 56.31 & 61.45 \\
        \bf FSUIE-base & BERT-base & 91.78 & 58.99 & 71.82 & 96.79 & 57.69 & 72.29  & 91.10 & 75.87 & 82.79  \\
        \bf FSUIE-large & BERT-large &89.01 &61.13 &\textbf{72.48} &98.84 &59.34 &\textbf{74.16} &92.02 &78.23 &\textbf{84.57} \\
        \bottomrule
    \end{tabular}
    }
    \caption{RE experimental results on ACE04, ACE05, and ADE datasets.}
    \label{tab:re_results}
\end{table*}

\subsection{Results on NER tasks}

We report the results of NER task in Table \ref{tab:ner_results}. By comparing the results of our baseline UIE-base with other methods, it can be seen that UIE-base has achieved comparable results compared to other methods that use the same BERT-base architecture. It serves as a strong baseline to visually demonstrate enhancements made by FSL and FSA. By introducing FSL and FSA, our FSUIE-base achieves significant performance improvements over the UIE-base that does not have fuzzy span mechanism (+1.15, +1.59, +1.99 F1 scores). 
Our proposed FSUIE model shows the most significant improvement on the ADE dataset. This is primarily due to the smaller scale of training datasets in the ADE dataset, which allows the model to easily learn generalized fuzzy span-aware representations. This demonstrates the superiority of the FSUIE model.

FSL and FSA enable the model to reduce over-dependence on label span boundaries and learn span-aware representations. When compared to existing NER models, FSUIE achieves new state-of-the-art performance on the ADE dataset even with the BERT-base backbone. FSUIE-large even achieve significant improvement (+1.42) on FSUIE-base. 
FSUIE-large also achieves comparable results on the ACE04 and ACE05 datasets, even when compared to models using stronger pre-trained language models such as ALBERT-xxlarge.
Furthermore, our FSUIE demonstrates an advantage in terms of its structure prediction compared to the generative UIE model. As it does not require the generation of complex IE linearized sequences, our FSUIE-base, which only uses BERT-base as its backbone, outperforms the generative UIE model that uses T5-v1.1-large on the ACE05 dataset.

\subsection{Results on RE tasks}

In Table \ref{tab:re_results}, we present the results of the RE tasks. Compared to the baseline, UIE-base, which does not incorporate fuzzy span mechanism, our proposed FSUIE-base, which incorporates FSL and FSA, also achieves a significant improvement on the RE task using same backtone. Furthermore, when compared to the Table-Sequence Encoder approach~\cite{wang-lu-2020-two}, our method learns label span boundary distribution and span-aware representations, resulting in optimal or competitive results on the RE task even with FSUIE-base, despite using a simpler structure and smaller PLM backbones.

Compared to span-based IE models, our method outperforms the traditional joint extraction model by performing a two stage span extraction and introducing the fuzzy span mechanism. Specifically, on the ADE dataset, our method performs better than joint extraction methods using Bio-BERT, a domain-specific pre-trained language model on biomedical corpus, even using BERT-base as the pre-trained language model. This demonstrates that the fuzzy span mechanism we introduced can extract general information from the data, giving the model stronger information extraction capabilities, rather than simply fitting the data.

Compare to generative UIE models, our span-based FSUIE reflects the reality of the structure of IE task and does not require additional sequence generation structures, achieving higher results with less parameters even with FSUIE-base.
Compared to models that perform relation extraction using a pipeline approach, like PL-Marker, our FSUIE improves performance in both stages of the pipeline by introducing FSL and FSA. As a result, it results in an overall improvement in relation extraction. Additionally, our model achieves new state-of-the-art results on ACE04 and ADE datasets,even using only BERT-base as the backbone, and on ACE05 dataset with FSUIE-large, compared to other models that use more complex structures. This demonstrates the model's ability to effectively extract information through our proposed method.

\begin{table*}[t]
    \resizebox{\textwidth}{!}{
    \begin{tabular}{lccccccccccccc}
    \toprule
    \multirow{2}{*}{\textbf{Models}} & \multirow{2}{*}{\textbf{PLM}} & \multicolumn{3}{c}{\textbf{14lap}} & \multicolumn{3}{c}{\textbf{14res}} & \multicolumn{3}{c}{\textbf{15res}} & \multicolumn{3}{c}{\textbf{16res}} \\
    \cmidrule(lr){3-5} \cmidrule(lr){6-8} \cmidrule(lr){9-11} \cmidrule(lr){12-14} & & P & R & F$_1$     & P & R & F$_1$     & P & R & F$_1$   & P & R & F$_1$  \\
    \midrule
    JET~\cite{xu-etal-2020-position} & BERT-base  & 55.39 & 47.33 & 51.04 & 70.56 & 55.94      & 62.40 & 64.45 & 51.96 & 57.53 & 70.42 & 58.37 & 63.83 \\
    Dual-MRC~\cite{DBLP:conf/aaai/MaoSYC21} & BERT-base & 57.39 & 53.88 & 55.58 & 71.55 & 69.14 & 70.32  & 63.78 & 51.87 & 57.21 & 68.60 & 66.24 & 67.40  \\
    ASTE-RL~\cite{DBLP:conf/aaai/MaoSYC21} & BERT-base  & 64.80 & 54.99 & 59.50 & 70.60  & 68.65 & 69.61 & 65.45 & 60.29 & 62.72 & 67.21 &69.69 & 68.41 \\
    GAS~\cite{zhang-etal-2021-towards-generative} & BERT-base & - & - & 60.78  & - & -  & 72.16  & - & - & 62.10  & - & - & 70.10 \\
    Span-ASTE~\cite{xu-etal-2021-learning} & BERT-base & 63.44 & 55.84 & 59.38 & 72.89 & 70.89 & 71.85 & 62.18 & 64.45 & 63.27 & 69.45 & 71.17 & 70.26 \\
    SBN~\cite{chen-2022-SBN} & BERT-base & 65.68 & 59.88 & 62.65 & 76.36  & 72.43 & 74.34      & 69.93 & 60.41 & 64.82 & 71.59 &72.57  & 72.08  \\
    \midrule
    Generative UIE (SEL only)~\cite{lu-etal-2022-unified} & T5-v1.1-large & - & - & 63.15  & - & - & 73.78  & - & -& 66.10  & -  & - & 73.87 \\
    Generative UIE~\cite{lu-etal-2022-unified} & T5-v1.1-large & - & - & 63.88  & - & - & \textbf{74.52} & - & - & 67.15 & - & - & 75.07 \\
    \midrule
    UIE-base  & BERT-base &65.21 &57.64  &61.19 & 75.32 & 71.53 & 73.38 &71.11  &65.98 &68.45  &74.84  &70.04  &72.36  \\
    \bf FSUIE-base & BERT-base & 69.49  & 62.06 & \textbf{65.56} & 76.22 & 72.23 & 74.17 & 72.71 & 68.66 & \textbf{70.63}& 77.98  & 73.74 & \textbf{75.80} \\
    \bottomrule
    \end{tabular}}
    \caption{ASTE experimental results on ASTE-DATA-V2 datasets (14lap, 14res, 15res, and 16res).}
    \label{tab:astev2_results}
\end{table*}

\subsection{Results on ASTE tasks}

In Table~\ref{tab:astev2_results}, we present the results of our experiments on the ASTE task. Due to the small scale of the ASTE-Data-V2, FSUIE-large is not needed to achieve better results, and this section only uses FSUIE-base for comparison. It can be seen that by introducing the fuzzy span mechanism, our FSUIE model significantly improves ASTE performance compared to the baseline UIE-base. This also demonstrates the effectiveness and generalization ability of FSUIE in IE tasks. Additionally, our FSUIE-base model achieves state-of-the-art results on three datasets (14lap, 15res, 16res) and demonstrates competitive performance on the 14res dataset. This indicates that the fuzzy span mechanism is effective in improving the model's ability to exploit and extract information, as well as its performance on specific tasks without increasing model parameters.

Furthermore, our FSUIE model has a relatively simple architecture, compared to other models, which shows that FSUIE is able to improve performance without the need for complex structures. The gap in performance between UIE models and other models can be attributed, in part, to the advantage of UIE pre-training, which is further enhanced by our proposed fuzzy span mechanism.
Compared to models that decompose the ASTE task into two subtasks of opinion recognition and sentiment classification, and use separate models to handle each, our FSUIE model achieved better performance using a unified model architecture.

For ASTE, span-based UIE models, as opposed to generative UIE models, can leverage the complete semantic information of the predicted aspect span to assist in extracting opinions and sentiments. The fuzzy span mechanism enhances the model's ability to exploit the semantic information within the fuzzy span, where possible opinions and sentiments reside, while ensuring span-aware representation learning, resulting in significant improvements.Furthermore, FSUIE is a reaction to the real structure of IEtask, avoiding the extra parameters that sequence generation structures bring, and therefore outperforms generative UIE models with fewer parameters.

We notice that FSUIE improves relatively less on the RE task compared to the ASTE task. In the RE task, the model has to learn different entities, different types of relationships, and binary matching skills. In contrast, in ASTE tasks, the model only needs to learn different entities, two relationships that differ significantly in semantics (opinion and sentiment), and ternary pairing tips. From this perspective, RE tasks are more challenging than ASTE tasks.

\subsection{Results on Low-resource Settings}

\begin{table}[]
    \resizebox{\columnwidth}{!}{
    \begin{tabular}{lllll}
    \toprule
    \textbf{Entity F1} & \textbf{1\%} & \textbf{5\%} & \textbf{25\%} & \textbf{100\%} \\
    \midrule
    UIE-base  & 63.47  & 72.98 & 83.08  & 84.63 \\
    \bf FSUIE-base  & 70.09 & 77.20  & 83.49  & 85.22 \\
    \midrule
    \textbf{Relation Strict F1}   & \textbf{1\%} & \textbf{5\%} & \textbf{25\%} & \textbf{100\%} \\
    \midrule
    UIE-base  & 6.68  & 45.55 & 64.43 & 66.72 \\
    \bf FSUIE-base     & 9.73  & 53.44  & 66.08 & 71.82 \\
    \midrule
    \textbf{Sentiment Triplet F1} & \textbf{1\%} & \textbf{5\%} & \textbf{25\%} & \textbf{100\%} \\
    \midrule
    UIE-base & 45.66 & 63.12 & 73.73 & 73.38 \\
    \bf FSUIE-base & 46.87 & 63.79 & 74.27 & 74.17 \\    
    \bottomrule
    \end{tabular}}
    \caption{Experimental results on low-resource settings.}
    \label{tab:low_resource_results}
\end{table}

To demonstrate the robustness of our proposed FSUIE method in low-resource scenarios, we conducted experiments using a reduced amount of training data on ACE04 for NER and RE tasks, and 14res for ASTE task. Specifically, we created three subsets of the original training data at 1\%, 5\%, and 25\% of the original size. In each low-resource experiment, we trained the model for 200 epochs instead of 50 epochs. The results of these experiments were compared between FSUIE-base and UIE-base and are presented in Table~\ref{tab:low_resource_results}.

The results of the low-resource experiments further confirm the superior performance of FSUIE over UIE in handling low-resource scenarios. With only a small fraction of the original training data, FSUIE is still able to achieve competitive or even better performance than UIE. This demonstrates the robustness and generalization ability of FSUIE in dealing with limited data. Overall, the results of the low-resource experiments validate the ability of FSUIE to effectively handle low-resource scenarios and extract rich information through limited data.

We also found that the model both performed better on NER and ASTE taks than on RE task under low-resource settings. This is because NER and ASTE tasks are simpler than RE, so less data can bring better learning performance. Additionally, we noticed a small performance decrease in the ASTE task for the 100\% set compared to the 25\% set. This change may be due to the fact that the training data is unbalanced, and reducing the training size can alleviate this phenomenon.

\subsection{Ablation Study}

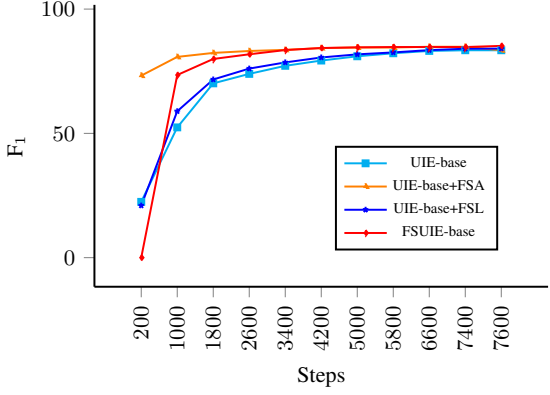
\begin{figure}[t ]
	\centering
	\scalebox{.93}{
	    \setlength{\abovecaptionskip}{0pt}
		\pgfplotsset{height=5.6cm,width=9cm,compat=1.16,every axis/.append style={thick},every axis legend/.append style={at={(0.89,0.5)}},legend columns=1}
		\begin{tikzpicture}
		\tikzset{every node}=[font=\small]
        \begin{axis}[width=8cm,enlargelimits=0.13, tick align=outside, xticklabels={$200$, $1000$,$1800$, $2600$, $3400$, $4200$, $5000$, $5800$, $6600$, $7400$, $7600$},
                    xticklabel style={
                    rotate=90,
                    },
                    axis x line*=bottom,
                    xtick={0,1,2,3,4,5,6,7,8,9,10},
                    ylabel={F$_1$},
                    axis y line*=left,
                    ylabel style={align=left},xlabel={Steps},ymax=90,font=\small]
			\addplot+ [sharp plot,mark=square*,mark size=1.2pt,mark options={solid,mark color=cyan}, color=cyan] coordinates
			{ (0,22.452)(1,52.411)(2,70.122)(3,73.932)(4,77.176)(5,79.296)(6,81.01)(7,82.227)(8,83.165)(9,83.437)(10,83.437) };
			\addlegendentry{\tiny UIE-base};\label{plot_a}
			\addplot+ [sharp plot,mark=triangle*,mark size=1.2pt,mark options={dashed,mark color=orange}, color=orange] coordinates
			{ (0,73.228)(1,80.765)(2,82.365)(3,83.146)(4,83.596)(5,84.313)(6,84.55)(7,84.55)(8,84.845)(9,84.845)(10,83.437) };
			\addlegendentry{\tiny UIE-base+FSA};\label{plot_b}
            \addplot+ [sharp plot,mark=star,mark size=1.2pt,mark options={dashed,mark color=blue}, color=blue] coordinates
			{ (0,20.947)(1,58.9)(2,71.677)(3,76.062)(4,78.544)(5,80.532)(6,81.789)(7,82.571)(8,83.491)(9,84.053)(10,84.116) };
			\addlegendentry{\tiny UIE-base+FSL};\label{plot_c}
            \addplot+ [sharp plot,mark=diamond*,mark size=1.2pt,mark options={dashed,mark color=red}, color=red] coordinates
			{ (0,0)(1,73.483)(2,79.932)(3,81.816)(4,83.514)(5,84.344)(6,84.6)(7,84.735)(8,84.735)(9,84.735)(10,85.196) };
			\addlegendentry{\tiny FSUIE-base};\label{plot_d}
        \end{axis}
		\end{tikzpicture}}
	\caption{NER performance of different models on ACE04 test set.}\label{fig:vary}
\end{figure}

Since FSUIE has been verified to make more effective use of the information in the training set, in order to verify this, we verify it from the perspective of the model training process. Specifically, we recorded the effects of baseline UIE-base, UIE-base+FSL, UIE-base+FSA and full model FSUIE-base on different training steps on the NER ACE04 test set, and the results are shown in Figure \ref{fig:vary}.

We noticed that the models with FSA have a significantly faster convergence speed, indicating that by learning span-aware representations, which are closer to the span prediction goal, the span learning process becomes more easy and efficient. With FSA, the model can focus its attention on the necessary positions and capture the possible span within a given sequence. While for FSL, it have a similar convergence trend with the baseline, thus may not improve the convergence speed.

\begin{table}[!ht]
    \centering
    \small
    \begin{tabular}{llll}
        \toprule
        \textbf{Models} & \textbf{P} & \textbf{R} & \textbf{F1 } \\ 
        \midrule
        UIE-base & 87.85 & 91.56 & 89.67  \\ 
        UIE-base + FSL & 89.61 & 90.58 & 90.09  \\ 
        UIE-base + FSA & 89.21 & 90.09 & 89.65  \\ 
        FSUIE-base & 91.17 & 92.17 & 92.49  \\ 
        \bottomrule
    \end{tabular}
    \caption{Ablation study of FSL and FSA on NER task using ADE dataset.}\label{tab:ablation_results}
\end{table}

To further investigate the contribution of FSL and FSA to the improvement of model performance, we conduct ablation experiments on the NER task using the ADE dataset. The specific experimental results are shown in Table~\ref{tab:ablation_results}. It can be seen that the introduction of FSL alone can improve model performance individually. When using FSA alone, the performance of the model drops slightly. However, when both FSL and FSA are used together, the model is significantly enhanced.

From our perspectives, the separate introduction of FSA makes the model focus on specific parts of the sequence rather than global representation, resulting in a loss of information from text outside the span. This may explain the slight drop in performance when using UIE+FSA. However, this also demonstrates that in the IE task, sequence information outside a specific span has a very limited impact on the results. The introduction of FSL alleviates the model's over-dependence on label span boundaries, allowing the model to extract more information, resulting in an improvement in both settings. When FSA and FSL operate simultaneously, the model extracts more information from the text and FSA guides the model to filter the more critical information from the richer information, resulting in the most substantial improvement.

\subsection{Visualization of FSA}

\begin{figure}[t]
    
    \centering    
    \includegraphics[width=\columnwidth]{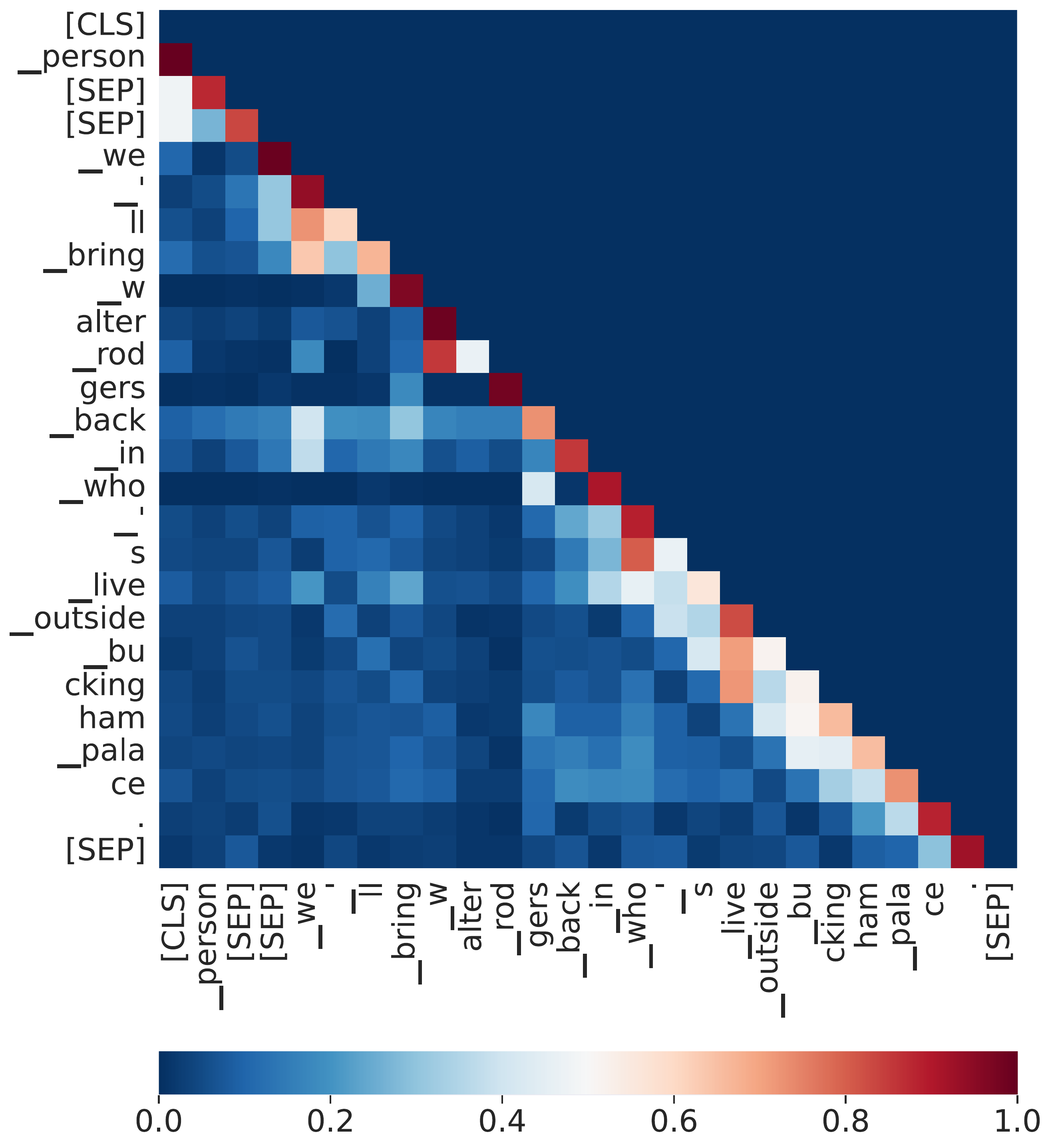}
    \vspace{-10pt}
    \caption{Illustration of attention scores distribution in FSA layer. The extracting target are "walter rodgers" and "who".}     
    \label{fig:fsa_weights}     
\end{figure}

To further examine the effectiveness of the fuzzy span mechanism, we visualized the attention distribution of the FSA layer in FSUIE-large as shown in Figure~\ref{fig:fsa_weights}. 
It should be noted that FSA is only placed at the top layer for constructing span-aware representation and does not participate in the encoding process, thus only affects span decisions rather than the representation of tokens in the sequence. 

The attention distribution indicates that, for a given input text, each token in the final encoding sequence tends to focus on semantic information within a limited range of preceding tokens rather than on the global representation of the input text. This aligns with our expectation for the design of the fuzzy span mechanism and confirms that fuzzy span mechanism does indeed guide the model for appropriate attention distribution for IE tasks.

\section{Related Work}

\paragraph{Universal Model}

Building universal model structures for a wide range of NLP tasks has been a hot research area in recent years. 
The focus is on building model structures that can be adapted to different sources of data, different types of labels, different languages, and different tasks. 
Several universal models have been proposed, such as models learning deep contextualized word representations~\cite{peters-etal-2018-deep, devlin-etal-2019-bert}, event extraction models that can predict different labels universally~\cite{lu-etal-2021-text2event}, models that can handle multiple languages~\cite{DBLP:journals/corr/abs-1907-05019, aharoni-etal-2019-massively, conneau-etal-2020-unsupervised}, a universal fine-tuned approach to transfer learning~\cite{howard-ruder-2018-universal}, models that learning syntactic dependency structure over many typologically different languages~\cite{li-etal-2018-joint-learning,sun2020cross} 
and models that can universally model various IE tasks in a unified text-to-structure framework~\cite{lu-etal-2022-unified}. This paper builds upon the UIE by incorporating the fuzzy span mechanism to improve IE performance.

\paragraph{Information Extraction}

IE is the task of extracting structured information from unstructured text data. This includes NER, RE, ASTE, Event Extraction (EE), Aspect-Based Sentiment Analysis (ABSA), etc. Research has proposed numerous approaches for IE, such as rule-based~\cite{appelt-onyshkevych-1998-common}, machine learning~\cite{DBLP:conf/cicling/Tellez-ValeroMP05, kolya-etal-2010-supervised}, deep learning~\cite{qin-etal-2018-robust}, active learning~\cite{radmard-etal-2021-subsequence}, and logic fusion~\cite{DBLP:conf/aaai/WangP20}.
There are still many task-specific models being proposed, based on previous approaches and structures, e.g., NER~\cite{li-etal-2022-unsupervised-multiple, zhu-li-2022-boundary,DBLP:conf/coling/YangLZ22,DBLP:journals/corr/abs-1909-01065}; RE~\cite{nan-etal-2020-reasoning, lai-etal-2021-joint}, ABSA~\cite{jing-etal-2021-seeking}; and ASTE~\cite{xu-etal-2021-learning, xu-etal-2020-position}. 

More related work about sparse attention please refer to Appendix B.

\section{Conclusion}

In this paper, we proposed the Fuzzy Span Universal Information Extraction (FSUIE) framework, an improvement for Universal Information Extraction. To make use of boundary information in the training data and learn a decision-closer span-aware representation, we proposed a fuzzy span loss and fuzzy span attention. Extensive experiments on several main IE tasks show that our FSUIE has a significant improvement compared to the UIE baseline, and achieves state-of-the-art results on ADE NER datasets, ACE04 RE, ACE05 RE and ADE RE datasets and four ASTE datasets. The experiments also reveal FSUIE's fast convergence and good generality in low-resource settings. All the results demonstrate the effectiveness and generalizability of our FSUIE in information extraction.

\section{Limitations}

This paper are based on the assumption that Universal Information Extraction (UIE) models have limitations, particularly with regards to over-reliance on label span boundaries and inflexible attention span length. Therefore, the proposed framework may be computationally and spatially expensive as it requires a more complex attention mechanism and additional computing power for training. Nevertheless, this limitation of the span-based UIE model can be overlooked in comparison to that of the generative UIE model, which uses a stronger language model. Additionally, the probability density functions explored in FSL are limited; thus, further research is needed to develop a more targeted strategy for adjusting the correct information distribution.

\bibliography{custom}

\begin{thebibliography}{45}
\expandafter\ifx\csname natexlab\endcsname\relax\def\natexlab#1{#1}\fi

\bibitem[{Aharoni et~al.(2019)Aharoni, Johnson, and
  Firat}]{aharoni-etal-2019-massively}
Roee Aharoni, Melvin Johnson, and Orhan Firat. 2019.
\newblock \href {https://doi.org/10.18653/v1/N19-1388} {Massively multilingual
  neural machine translation}.
\newblock In \emph{Proceedings of the 2019 Conference of the North {A}merican
  Chapter of the Association for Computational Linguistics: Human Language
  Technologies, Volume 1 (Long and Short Papers)}, pages 3874--3884,
  Minneapolis, Minnesota. Association for Computational Linguistics.

\bibitem[{Appelt and Onyshkevych(1998)}]{appelt-onyshkevych-1998-common}
Douglas~E. Appelt and Boyan Onyshkevych. 1998.
\newblock \href {https://doi.org/10.3115/1119089.1119095} {The common pattern
  specification language}.
\newblock In \emph{TIPSTER TEXT PROGRAM PHASE III: Proceedings of a Workshop
  held at Baltimore, {M}aryland, October 13-15, 1998}, pages 23--30, Baltimore,
  Maryland, USA. Association for Computational Linguistics.

\bibitem[{Arivazhagan et~al.(2019)Arivazhagan, Bapna, Firat, Lepikhin, Johnson,
  Krikun, Chen, Cao, Foster, Cherry, Macherey, Chen, and
  Wu}]{DBLP:journals/corr/abs-1907-05019}
Naveen Arivazhagan, Ankur Bapna, Orhan Firat, Dmitry Lepikhin, Melvin Johnson,
  Maxim Krikun, Mia~Xu Chen, Yuan Cao, George~F. Foster, Colin Cherry, Wolfgang
  Macherey, Zhifeng Chen, and Yonghui Wu. 2019.
\newblock \href {http://arxiv.org/abs/1907.05019} {Massively multilingual
  neural machine translation in the wild: Findings and challenges}.
\newblock \emph{CoRR}, abs/1907.05019.

\bibitem[{Chen et~al.(2022)Chen, Chen, Sun, and Zhang}]{chen-2022-SBN}
Yuqi Chen, Keming Chen, Xian Sun, and Zequn Zhang. 2022.
\newblock A span-level bidirectional network for aspect sentiment triplet
  extraction.
\newblock In \emph{Proceedings of the 2022 Conference on Empirical Methods in
  Natural Language Processing (EMNLP)}.

\bibitem[{Child et~al.(2019)Child, Gray, Radford, and
  Sutskever}]{DBLP:journals/corr/abs-1904-10509}
Rewon Child, Scott Gray, Alec Radford, and Ilya Sutskever. 2019.
\newblock \href {http://arxiv.org/abs/1904.10509} {Generating long sequences
  with sparse transformers}.
\newblock \emph{CoRR}, abs/1904.10509.

\bibitem[{Conneau et~al.(2020)Conneau, Khandelwal, Goyal, Chaudhary, Wenzek,
  Guzm{\'a}n, Grave, Ott, Zettlemoyer, and
  Stoyanov}]{conneau-etal-2020-unsupervised}
Alexis Conneau, Kartikay Khandelwal, Naman Goyal, Vishrav Chaudhary, Guillaume
  Wenzek, Francisco Guzm{\'a}n, Edouard Grave, Myle Ott, Luke Zettlemoyer, and
  Veselin Stoyanov. 2020.
\newblock \href {https://doi.org/10.18653/v1/2020.acl-main.747} {Unsupervised
  cross-lingual representation learning at scale}.
\newblock In \emph{Proceedings of the 58th Annual Meeting of the Association
  for Computational Linguistics}, pages 8440--8451, Online. Association for
  Computational Linguistics.

\bibitem[{Devlin et~al.(2019)Devlin, Chang, Lee, and
  Toutanova}]{devlin-etal-2019-bert}
Jacob Devlin, Ming-Wei Chang, Kenton Lee, and Kristina Toutanova. 2019.
\newblock \href {https://doi.org/10.18653/v1/N19-1423} {{BERT}: Pre-training of
  deep bidirectional transformers for language understanding}.
\newblock In \emph{Proceedings of the 2019 Conference of the North {A}merican
  Chapter of the Association for Computational Linguistics: Human Language
  Technologies, Volume 1 (Long and Short Papers)}, pages 4171--4186,
  Minneapolis, Minnesota. Association for Computational Linguistics.

\bibitem[{Eberts and Ulges(2019)}]{DBLP:journals/corr/abs-1909-07755}
Markus Eberts and Adrian Ulges. 2019.
\newblock \href {http://arxiv.org/abs/1909.07755} {Span-based joint entity and
  relation extraction with transformer pre-training}.
\newblock \emph{CoRR}, abs/1909.07755.

\bibitem[{Gurulingappa et~al.(2012)Gurulingappa, Rajput, Roberts, Fluck,
  Hofmann{-}Apitius, and Toldo}]{DBLP:journals/jbi/GurulingappaRRFHT12}
Harsha Gurulingappa, Abdul~Mateen Rajput, Angus Roberts, Juliane Fluck, Martin
  Hofmann{-}Apitius, and Luca Toldo. 2012.
\newblock \href {https://doi.org/10.1016/j.jbi.2012.04.008} {Development of a
  benchmark corpus to support the automatic extraction of drug-related adverse
  effects from medical case reports}.
\newblock \emph{J. Biomed. Informatics}, 45(5):885--892.

\bibitem[{He et~al.(2022)He, Mao, Gong, Cambria, and Li}]{he2022jcbie}
Kai He, Rui Mao, Tieliang Gong, Erik Cambria, and Chen Li. 2022.
\newblock Jcbie: a joint continual learning neural network for biomedical
  information extraction.
\newblock \emph{BMC bioinformatics}, 23(1):1--20.

\bibitem[{Howard and Ruder(2018)}]{howard-ruder-2018-universal}
Jeremy Howard and Sebastian Ruder. 2018.
\newblock \href {https://doi.org/10.18653/v1/P18-1031} {Universal language
  model fine-tuning for text classification}.
\newblock In \emph{Proceedings of the 56th Annual Meeting of the Association
  for Computational Linguistics (Volume 1: Long Papers)}, pages 328--339,
  Melbourne, Australia. Association for Computational Linguistics.

\bibitem[{Jiang et~al.(2021)Jiang, Wang, Chen, Zhang, and
  Karlsson}]{DBLP:journals/corr/abs-2107-09429}
Huiqiang Jiang, Guoxin Wang, Weile Chen, Chengxi Zhang, and B{\"{o}}rje~F.
  Karlsson. 2021.
\newblock \href {http://arxiv.org/abs/2107.09429} {Boningknife: Joint entity
  mention detection and typing for nested {NER} via prior boundary knowledge}.
\newblock \emph{CoRR}, abs/2107.09429.

\bibitem[{Jing et~al.(2021)Jing, Li, Zhao, and Jiang}]{jing-etal-2021-seeking}
Hongjiang Jing, Zuchao Li, Hai Zhao, and Shu Jiang. 2021.
\newblock \href {https://doi.org/10.18653/v1/2021.emnlp-main.318} {Seeking
  common but distinguishing difference, a joint aspect-based sentiment analysis
  model}.
\newblock In \emph{Proceedings of the 2021 Conference on Empirical Methods in
  Natural Language Processing}, pages 3910--3922, Online and Punta Cana,
  Dominican Republic. Association for Computational Linguistics.

\bibitem[{Kolya et~al.(2010)Kolya, Ekbal, and
  Bandyopadhyay}]{kolya-etal-2010-supervised}
Anup~Kumar Kolya, Asif Ekbal, and Sivaji Bandyopadhyay. 2010.
\newblock \href {https://aclanthology.org/Y10-1051} {A supervised machine
  learning approach for event-event relation identification}.
\newblock In \emph{Proceedings of the 24th Pacific Asia Conference on Language,
  Information and Computation}, pages 447--454, Tohoku University, Sendai,
  Japan. Institute of Digital Enhancement of Cognitive Processing, Waseda
  University.

\bibitem[{Lai et~al.(2021)Lai, Ji, Zhai, and Tran}]{lai-etal-2021-joint}
Tuan Lai, Heng Ji, ChengXiang Zhai, and Quan~Hung Tran. 2021.
\newblock \href {https://doi.org/10.18653/v1/2021.acl-long.488} {Joint
  biomedical entity and relation extraction with knowledge-enhanced collective
  inference}.
\newblock In \emph{Proceedings of the 59th Annual Meeting of the Association
  for Computational Linguistics and the 11th International Joint Conference on
  Natural Language Processing (Volume 1: Long Papers)}, pages 6248--6260,
  Online. Association for Computational Linguistics.

\bibitem[{Li et~al.(2022)Li, Hu, Guo, Chen, Qin, and
  Zhang}]{li-etal-2022-unsupervised-multiple}
Zhuoran Li, Chunming Hu, Xiaohui Guo, Junfan Chen, Wenyi Qin, and Richong
  Zhang. 2022.
\newblock \href {https://doi.org/10.18653/v1/2022.acl-long.14} {An unsupervised
  multiple-task and multiple-teacher model for cross-lingual named entity
  recognition}.
\newblock In \emph{Proceedings of the 60th Annual Meeting of the Association
  for Computational Linguistics (Volume 1: Long Papers)}, pages 170--179,
  Dublin, Ireland. Association for Computational Linguistics.

\bibitem[{Li et~al.(2018)Li, He, Zhang, and Zhao}]{li-etal-2018-joint-learning}
Zuchao Li, Shexia He, Zhuosheng Zhang, and Hai Zhao. 2018.
\newblock \href {https://doi.org/10.18653/v1/K18-2006} {Joint learning of {POS}
  and dependencies for multilingual {U}niversal {D}ependency parsing}.
\newblock In \emph{Proceedings of the {C}o{NLL} 2018 Shared Task: Multilingual
  Parsing from Raw Text to Universal Dependencies}, pages 65--73, Brussels,
  Belgium. Association for Computational Linguistics.

\bibitem[{Lu et~al.(2021)Lu, Lin, Xu, Han, Tang, Li, Sun, Liao, and
  Chen}]{lu-etal-2021-text2event}
Yaojie Lu, Hongyu Lin, Jin Xu, Xianpei Han, Jialong Tang, Annan Li, Le~Sun,
  Meng Liao, and Shaoyi Chen. 2021.
\newblock \href {https://doi.org/10.18653/v1/2021.acl-long.217}
  {{T}ext2{E}vent: Controllable sequence-to-structure generation for end-to-end
  event extraction}.
\newblock In \emph{Proceedings of the 59th Annual Meeting of the Association
  for Computational Linguistics and the 11th International Joint Conference on
  Natural Language Processing (Volume 1: Long Papers)}, pages 2795--2806,
  Online. Association for Computational Linguistics.

\bibitem[{Lu et~al.(2022)Lu, Liu, Dai, Xiao, Lin, Han, Sun, and
  Wu}]{lu-etal-2022-unified}
Yaojie Lu, Qing Liu, Dai Dai, Xinyan Xiao, Hongyu Lin, Xianpei Han, Le~Sun, and
  Hua Wu. 2022.
\newblock \href {https://doi.org/10.18653/v1/2022.acl-long.395} {Unified
  structure generation for universal information extraction}.
\newblock In \emph{Proceedings of the 60th Annual Meeting of the Association
  for Computational Linguistics (Volume 1: Long Papers)}, pages 5755--5772,
  Dublin, Ireland. Association for Computational Linguistics.

\bibitem[{Ma et~al.(2020)Ma, Hiraoka, and
  Okazaki}]{DBLP:journals/corr/abs-2010-07522}
Youmi Ma, Tatsuya Hiraoka, and Naoaki Okazaki. 2020.
\newblock \href {http://arxiv.org/abs/2010.07522} {Named entity recognition and
  relation extraction using enhanced table filling by contextualized
  representations}.
\newblock \emph{CoRR}, abs/2010.07522.

\bibitem[{Mao et~al.(2021)Mao, Shen, Yu, and Cai}]{DBLP:conf/aaai/MaoSYC21}
Yue Mao, Yi~Shen, Chao Yu, and Longjun Cai. 2021.
\newblock \href {https://ojs.aaai.org/index.php/AAAI/article/view/17597} {A
  joint training dual-mrc framework for aspect based sentiment analysis}.
\newblock In \emph{Thirty-Fifth {AAAI} Conference on Artificial Intelligence,
  {AAAI} 2021, Thirty-Third Conference on Innovative Applications of Artificial
  Intelligence, {IAAI} 2021, The Eleventh Symposium on Educational Advances in
  Artificial Intelligence, {EAAI} 2021, Virtual Event, February 2-9, 2021},
  pages 13543--13551. {AAAI} Press.

\bibitem[{Nan et~al.(2020)Nan, Guo, Sekulic, and Lu}]{nan-etal-2020-reasoning}
Guoshun Nan, Zhijiang Guo, Ivan Sekulic, and Wei Lu. 2020.
\newblock \href {https://doi.org/10.18653/v1/2020.acl-main.141} {Reasoning with
  latent structure refinement for document-level relation extraction}.
\newblock In \emph{Proceedings of the 58th Annual Meeting of the Association
  for Computational Linguistics}, pages 1546--1557, Online. Association for
  Computational Linguistics.

\bibitem[{Peters et~al.(2018)Peters, Neumann, Iyyer, Gardner, Clark, Lee, and
  Zettlemoyer}]{peters-etal-2018-deep}
Matthew~E. Peters, Mark Neumann, Mohit Iyyer, Matt Gardner, Christopher Clark,
  Kenton Lee, and Luke Zettlemoyer. 2018.
\newblock \href {https://doi.org/10.18653/v1/N18-1202} {Deep contextualized
  word representations}.
\newblock In \emph{Proceedings of the 2018 Conference of the North {A}merican
  Chapter of the Association for Computational Linguistics: Human Language
  Technologies, Volume 1 (Long Papers)}, pages 2227--2237, New Orleans,
  Louisiana. Association for Computational Linguistics.

\bibitem[{Qin et~al.(2018)Qin, Xu, and Wang}]{qin-etal-2018-robust}
Pengda Qin, Weiran Xu, and William~Yang Wang. 2018.
\newblock \href {https://doi.org/10.18653/v1/P18-1199} {Robust distant
  supervision relation extraction via deep reinforcement learning}.
\newblock In \emph{Proceedings of the 56th Annual Meeting of the Association
  for Computational Linguistics (Volume 1: Long Papers)}, pages 2137--2147,
  Melbourne, Australia. Association for Computational Linguistics.

\bibitem[{Radmard et~al.(2021)Radmard, Fathullah, and
  Lipani}]{radmard-etal-2021-subsequence}
Puria Radmard, Yassir Fathullah, and Aldo Lipani. 2021.
\newblock \href {https://doi.org/10.18653/v1/2021.acl-long.332} {Subsequence
  based deep active learning for named entity recognition}.
\newblock In \emph{Proceedings of the 59th Annual Meeting of the Association
  for Computational Linguistics and the 11th International Joint Conference on
  Natural Language Processing (Volume 1: Long Papers)}, pages 4310--4321,
  Online. Association for Computational Linguistics.

\bibitem[{Santosh et~al.(2021)Santosh, Chakraborty, Dutta, Sanyal, and
  Das}]{santosh2021joint}
TYSS Santosh, Prantika Chakraborty, Sudakshina Dutta, Debarshi~Kumar Sanyal,
  and Partha~Pratim Das. 2021.
\newblock Joint entity and relation extraction from scientific documents: Role
  of linguistic information and entity types.

\bibitem[{Su et~al.(2022)Su, Murtadha, Pan, Hou, Sun, Huang, Wen, and
  Liu}]{DBLP:journals/corr/abs-2208-03054}
Jianlin Su, Ahmed Murtadha, Shengfeng Pan, Jing Hou, Jun Sun, Wanwei Huang,
  Bo~Wen, and Yunfeng Liu. 2022.
\newblock \href {https://doi.org/10.48550/arXiv.2208.03054} {Global pointer:
  Novel efficient span-based approach for named entity recognition}.
\newblock \emph{CoRR}, abs/2208.03054.

\bibitem[{Sukhbaatar et~al.(2019)Sukhbaatar, Grave, Bojanowski, and
  Joulin}]{sukhbaatar-etal-2019-adaptive}
Sainbayar Sukhbaatar, Edouard Grave, Piotr Bojanowski, and Armand Joulin. 2019.
\newblock \href {https://doi.org/10.18653/v1/P19-1032} {Adaptive attention span
  in transformers}.
\newblock In \emph{Proceedings of the 57th Annual Meeting of the Association
  for Computational Linguistics}, pages 331--335, Florence, Italy. Association
  for Computational Linguistics.

\bibitem[{Sun et~al.(2020)Sun, Li, and Zhao}]{sun2020cross}
Kailai Sun, Zuchao Li, and Hai Zhao. 2020.
\newblock Cross-lingual universal dependency parsing only from one monolingual
  treebank.
\newblock \emph{arXiv preprint arXiv:2012.13163}.

\bibitem[{Tan et~al.(2020)Tan, Qiu, Chen, Wang, and
  Huang}]{DBLP:conf/aaai/TanQCWH20}
Chuanqi Tan, Wei Qiu, Mosha Chen, Rui Wang, and Fei Huang. 2020.
\newblock \href {https://ojs.aaai.org/index.php/AAAI/article/view/6434}
  {Boundary enhanced neural span classification for nested named entity
  recognition}.
\newblock In \emph{The Thirty-Fourth {AAAI} Conference on Artificial
  Intelligence, {AAAI} 2020, The Thirty-Second Innovative Applications of
  Artificial Intelligence Conference, {IAAI} 2020, The Tenth {AAAI} Symposium
  on Educational Advances in Artificial Intelligence, {EAAI} 2020, New York,
  NY, USA, February 7-12, 2020}, pages 9016--9023. {AAAI} Press.

\bibitem[{T{\'{e}}llez{-}Valero et~al.(2005)T{\'{e}}llez{-}Valero,
  Montes{-}y{-}G{\'{o}}mez, and Pineda}]{DBLP:conf/cicling/Tellez-ValeroMP05}
Alberto T{\'{e}}llez{-}Valero, Manuel Montes{-}y{-}G{\'{o}}mez, and
  Luis~Villase{\~{n}}or Pineda. 2005.
\newblock \href {https://doi.org/10.1007/978-3-540-30586-6\_58} {A machine
  learning approach to information extraction}.
\newblock In \emph{Computational Linguistics and Intelligent Text Processing,
  6th International Conference, CICLing 2005, Mexico City, Mexico, February
  13-19, 2005, Proceedings}, volume 3406 of \emph{Lecture Notes in Computer
  Science}, pages 539--547. Springer.

\bibitem[{Vaswani et~al.(2017)Vaswani, Shazeer, Parmar, Uszkoreit, Jones,
  Gomez, Kaiser, and Polosukhin}]{DBLP:conf/nips/VaswaniSPUJGKP17}
Ashish Vaswani, Noam Shazeer, Niki Parmar, Jakob Uszkoreit, Llion Jones,
  Aidan~N. Gomez, Lukasz Kaiser, and Illia Polosukhin. 2017.
\newblock \href
  {https://proceedings.neurips.cc/paper/2017/hash/3f5ee243547dee91fbd053c1c4a845aa-Abstract.html}
  {Attention is all you need}.
\newblock In \emph{Advances in Neural Information Processing Systems 30: Annual
  Conference on Neural Information Processing Systems 2017, December 4-9, 2017,
  Long Beach, CA, {USA}}, pages 5998--6008.

\bibitem[{Wang and Lu(2020)}]{wang-lu-2020-two}
Jue Wang and Wei Lu. 2020.
\newblock \href {https://doi.org/10.18653/v1/2020.emnlp-main.133} {Two are
  better than one: Joint entity and relation extraction with table-sequence
  encoders}.
\newblock In \emph{Proceedings of the 2020 Conference on Empirical Methods in
  Natural Language Processing (EMNLP)}, pages 1706--1721, Online. Association
  for Computational Linguistics.

\bibitem[{Wang and Pan(2020)}]{DBLP:conf/aaai/WangP20}
Wenya Wang and Sinno~Jialin Pan. 2020.
\newblock \href {https://ojs.aaai.org/index.php/AAAI/article/view/6460}
  {Integrating deep learning with logic fusion for information extraction}.
\newblock In \emph{The Thirty-Fourth {AAAI} Conference on Artificial
  Intelligence, {AAAI} 2020, The Thirty-Second Innovative Applications of
  Artificial Intelligence Conference, {IAAI} 2020, The Tenth {AAAI} Symposium
  on Educational Advances in Artificial Intelligence, {EAAI} 2020, New York,
  NY, USA, February 7-12, 2020}, pages 9225--9232. {AAAI} Press.

\bibitem[{Xu et~al.(2021)Xu, Chia, and Bing}]{xu-etal-2021-learning}
Lu~Xu, Yew~Ken Chia, and Lidong Bing. 2021.
\newblock \href {https://doi.org/10.18653/v1/2021.acl-long.367} {Learning
  span-level interactions for aspect sentiment triplet extraction}.
\newblock In \emph{Proceedings of the 59th Annual Meeting of the Association
  for Computational Linguistics and the 11th International Joint Conference on
  Natural Language Processing (Volume 1: Long Papers)}, pages 4755--4766,
  Online. Association for Computational Linguistics.

\bibitem[{Xu et~al.(2020)Xu, Li, Lu, and Bing}]{xu-etal-2020-position}
Lu~Xu, Hao Li, Wei Lu, and Lidong Bing. 2020.
\newblock \href {https://doi.org/10.18653/v1/2020.emnlp-main.183}
  {Position-aware tagging for aspect sentiment triplet extraction}.
\newblock In \emph{Proceedings of the 2020 Conference on Empirical Methods in
  Natural Language Processing (EMNLP)}, pages 2339--2349, Online. Association
  for Computational Linguistics.

\bibitem[{Yan et~al.(2021)Yan, Gui, Dai, Guo, Zhang, and
  Qiu}]{yan-etal-2021-unified-generative}
Hang Yan, Tao Gui, Junqi Dai, Qipeng Guo, Zheng Zhang, and Xipeng Qiu. 2021.
\newblock \href {https://doi.org/10.18653/v1/2021.acl-long.451} {A unified
  generative framework for various {NER} subtasks}.
\newblock In \emph{Proceedings of the 59th Annual Meeting of the Association
  for Computational Linguistics and the 11th International Joint Conference on
  Natural Language Processing (Volume 1: Long Papers)}, pages 5808--5822,
  Online. Association for Computational Linguistics.

\bibitem[{Yang et~al.(2022)Yang, Li, and Zhao}]{DBLP:conf/coling/YangLZ22}
Yifei Yang, Zuchao Li, and Hai Zhao. 2022.
\newblock \href {https://aclanthology.org/2022.coling-1.218} {Nested named
  entity recognition as corpus aware holistic structure parsing}.
\newblock In \emph{Proceedings of the 29th International Conference on
  Computational Linguistics, {COLING} 2022, Gyeongju, Republic of Korea,
  October 12-17, 2022}, pages 2472--2482. International Committee on
  Computational Linguistics.

\bibitem[{Ye et~al.(2022)Ye, Lin, Li, and Sun}]{ye-etal-2022-packed}
Deming Ye, Yankai Lin, Peng Li, and Maosong Sun. 2022.
\newblock \href {https://doi.org/10.18653/v1/2022.acl-long.337} {Packed
  levitated marker for entity and relation extraction}.
\newblock In \emph{Proceedings of the 60th Annual Meeting of the Association
  for Computational Linguistics (Volume 1: Long Papers)}, pages 4904--4917,
  Dublin, Ireland. Association for Computational Linguistics.

\bibitem[{Yuan et~al.(2022)Yuan, Tan, Huang, and Huang}]{yuan-etal-2022-fusing}
Zheng Yuan, Chuanqi Tan, Songfang Huang, and Fei Huang. 2022.
\newblock \href {https://doi.org/10.18653/v1/2022.findings-acl.250} {Fusing
  heterogeneous factors with triaffine mechanism for nested named entity
  recognition}.
\newblock In \emph{Findings of the Association for Computational Linguistics:
  ACL 2022}, pages 3174--3186, Dublin, Ireland. Association for Computational
  Linguistics.

\bibitem[{Zaheer et~al.(2020)Zaheer, Guruganesh, Dubey, Ainslie, Alberti,
  Onta{\~{n}}{\'{o}}n, Pham, Ravula, Wang, Yang, and
  Ahmed}]{DBLP:conf/nips/ZaheerGDAAOPRWY20}
Manzil Zaheer, Guru Guruganesh, Kumar~Avinava Dubey, Joshua Ainslie, Chris
  Alberti, Santiago Onta{\~{n}}{\'{o}}n, Philip Pham, Anirudh Ravula, Qifan
  Wang, Li~Yang, and Amr Ahmed. 2020.
\newblock \href
  {https://proceedings.neurips.cc/paper/2020/hash/c8512d142a2d849725f31a9a7a361ab9-Abstract.html}
  {Big bird: Transformers for longer sequences}.
\newblock In \emph{Advances in Neural Information Processing Systems 33: Annual
  Conference on Neural Information Processing Systems 2020, NeurIPS 2020,
  December 6-12, 2020, virtual}.

\bibitem[{Zhang et~al.(2021)Zhang, Li, Deng, Bing, and
  Lam}]{zhang-etal-2021-towards-generative}
Wenxuan Zhang, Xin Li, Yang Deng, Lidong Bing, and Wai Lam. 2021.
\newblock \href {https://doi.org/10.18653/v1/2021.acl-short.64} {Towards
  generative aspect-based sentiment analysis}.
\newblock In \emph{Proceedings of the 59th Annual Meeting of the Association
  for Computational Linguistics and the 11th International Joint Conference on
  Natural Language Processing (Volume 2: Short Papers)}, pages 504--510,
  Online. Association for Computational Linguistics.

\bibitem[{Zhang et~al.(2019)Zhang, Tang, Li, and
  Zhao}]{DBLP:journals/corr/abs-1909-01065}
Zhuosheng Zhang, Bingjie Tang, Zuchao Li, and Hai Zhao. 2019.
\newblock \href {http://arxiv.org/abs/1909.01065} {Modeling named entity
  embedding distribution into hypersphere}.
\newblock \emph{CoRR}, abs/1909.01065.

\bibitem[{Zhu and Li(2022)}]{zhu-li-2022-boundary}
Enwei Zhu and Jinpeng Li. 2022.
\newblock \href {https://doi.org/10.18653/v1/2022.acl-long.490} {Boundary
  smoothing for named entity recognition}.
\newblock In \emph{Proceedings of the 60th Annual Meeting of the Association
  for Computational Linguistics (Volume 1: Long Papers)}, pages 7096--7108,
  Dublin, Ireland. Association for Computational Linguistics.

\bibitem[{Zhuang et~al.(2022)Zhuang, Zhang, and Tu}]{zhuang-etal-2022-long}
Yimeng Zhuang, Jing Zhang, and Mei Tu. 2022.
\newblock \href {https://doi.org/10.18653/v1/2022.acl-long.19} {Long-range
  sequence modeling with predictable sparse attention}.
\newblock In \emph{Proceedings of the 60th Annual Meeting of the Association
  for Computational Linguistics (Volume 1: Long Papers)}, pages 234--243,
  Dublin, Ireland. Association for Computational Linguistics.

\end{thebibliography}
\bibliographystyle{acl_natbib}

\clearpage
\section{Appendix}

\subsection{$g_{a}$ in FSA}

\begin{table}[!ht]
    \centering
    \small
    \begin{tabular}{llll}
        \toprule
        \textbf{} & \textbf{P} & \textbf{R} & \textbf{F1 } \\ 
        \midrule
        UIE-base & 87.85 & 91.56 & 89.67  \\ 
        FSUIE-base ($g'_{a}$) &89.84  &90.69  &90.26   \\ 
        FSUIE-base ($g''_{a}$) &90.93  &90.41  &90.67   \\ 
        FSUIE-base ($g^{l}_{a}$) & 91.17 & 92.17 & 92.49  \\ 
        \bottomrule
    \end{tabular}
    \caption{Performance of models using different $g_{a}$ in FSA}
    \label{tab:different_g_results}
\end{table}

In Table~\ref{tab:different_g_results}, we present the performance of models using various $g_{a}$ functions in the FSUIE technique on the ADE NER test set, where $g^{l}_{a}$ denotes the linear attenuated function employed in FSUIE. Compared to the UIE-base, which does not integrate the fuzzy span mechanism, all FSUIE-based models employing different $g_{a}$ functions obtain better results, thus illustrating the superiority of FSUIE. Regarding the different $g_{a}$ strategies, FSUIE-base ($g'_{a}$) shows minimal enhancement. This is likely because the fuzzy span of attention attenuation adequately reflects the real reading context and enables the model to take advantage of more abundant information within the boundary of the attention span. The best performance is achieved by FSUIE-base ($g^{l}_{a}$), which indicates that the attention should not decay too quickly at the boundary of the attention span, as evidenced by the results of $g''_{a}$.

\subsection{Related Work on Sparse Attention}

The high time and space complexity of Transformer ($O(n^2)$) is due to the fact that it needs to calculate the attention information between each step and all previous contexts. This makes it difficult for Transformer to scale in terms of sequence length. To address this issue, sparse attention was proposed~\cite{DBLP:journals/corr/abs-1904-10509}. This refers to attention mechanisms that focus on a small subset of the input elements, rather than processing the entire input sequence. This method allows attention to be more focused on the most contributing value factors, thus reducing memory and computing capacity requirements.

Based on the idea of sparse attention, various approaches have been proposed, such as an adaptive width-based attention learning mechanism and a dynamic attention mechanism that allows different heads to learn only the region of attention~\cite{sukhbaatar-etal-2019-adaptive}. \citet{ DBLP:conf/nips/ZaheerGDAAOPRWY20} proposed an $O(N)$ complexity model with three different sparse attentions. \citet{ zhuang-etal-2022-long} sought to make the sparse attention matrix predictable. This paper, however, based on adaptive span attention~\cite{sukhbaatar-etal-2019-adaptive} to establish a fuzzy span attention, which aims at learning a span-aware representation with the actual needs of information extraction tasks. Our approach differs from previous work in that we aim to obtain a fuzzy span of attention in the process of locating the target, rather than reducing computational and memory overhead.

\subsection{$L_{span}$ and $d$ in FSA}
\begin{table}[!ht]
    \centering
    \small
    \begin{tabular}{llll}
        \toprule
        \textbf{$d$} & \textbf{P} & \textbf{R} & \textbf{F1 } \\ 
        \midrule
        16 & 91.45 & 93.45 & 92.44  \\ 
        32 & 92.58 & 92.40 & 92.49  \\ 
        48 & 92.32 & 92.50 & 92.41  \\ 
        \bottomrule
    \end{tabular}
    \caption{Performance of FSUIE-base models with different $d$}
    \label{tab:different_d_results}
\end{table}

\begin{table}[!ht]
    \centering
    \small
    \begin{tabular}{llll}
        \toprule
        \textbf{$L_{span}$} & \textbf{P} & \textbf{R} & \textbf{F1 } \\ 
        \midrule
        16 & 91.65 & 93.83 & 92.73  \\ 
        30 & 92.58 & 92.40 & 92.49  \\ 
        48 & 92.25 & 92.69 & 92.47  \\ 
        \bottomrule
    \end{tabular}
    \caption{Performance of FSUIE-base models with different $L_{span}$}
    \label{tab:different_Lspan_results}
\end{table}

In Table~\ref{tab:different_d_results}, we present the performance of FSUIE-base models using various hyper-parameter $d$ on the ADE NER test set.
In Table~\ref{tab:different_Lspan_results}, we present the performance of FSUIE-base models using various hyper-parameter $L_{span}$ on the ADE NER test set.
The results demonstrate  that the model's performance is not significantly affected by the choice of these hyper-parameters.

\subsection{Single-Side and Both-Side Ambiguity in FSL}

\begin{table}[!ht]
    \centering
    \small
    \begin{tabular}{llll}
        \toprule
        \textbf{FSL strategies} & \textbf{P} & \textbf{R} & \textbf{F1 } \\ 
        \midrule
        both-side & 92.58 & 92.40 & 92.49  \\ 
        single-side & 91.99 & 92.69 & 92.34  \\ 
        \bottomrule
    \end{tabular}
    \caption{Performance of FSUIE-base models with different FSL strategies}
    \label{tab:different_side_results}
\end{table}

Actually, there may be cases of single-side ambiguity in the labeling of entity boundaries in the text. 
Therefore, we demonstrate FSUIE-base models' performance with different FSL strategies in Table~\ref{tab:different_side_results}, where "single-side" means applying FSL only on start boundary and "both-side" means applying FSL on both start and end boundary.
The results suggests that the influence of single-sided and both-sided fuzziness on the model's performance is limited, because not all head words are at the end or start, and FSL only performs limited left/right extrapolation on precise boundaries, without affecting the important information provided by the original boundary. For generalization purposes, we utilized both-sides fuzzy span in FSUIE.

\end{document}